# Travel Time, Distance and Costs Optimization for Paratransit Operations using Graph Convolutional Neural Network

Kelvin Kwakye, Younho Seong, Sun Yi

*Abstract*— The provision of paratransit services is one option to meet the transportation needs of Vulnerable Road Users (VRUs). Like any other means of transportation, paratransit has obstacles such as high operational costs and longer trip times. As a result, customers are dissatisfied, and paratransit operators have a low approval rating.

Researchers have undertaken various studies over the years to better understand the travel behaviors of paratransit customers and how they are operated. According to the findings of these research, paratransit operators confront the challenge of determining the optimal route for their trips in order to save travel time. Depending on the nature of the challenge, most research used different optimization techniques to solve these routing problems.

As a result, the goal of this study is to use Graph Convolutional Neural Networks (GCNs) to assist paratransit operators in researching various operational scenarios in a strategic setting in order to optimize routing, minimize operating costs and minimize their users' travel time. The study was carried out by using a randomized simulated dataset to help determine the decision to make in terms of fleet composition and capacity under different situations. For the various scenarios investigated, the GCN assisted in determining the minimum optimal gap.

*Index Terms*— Paratransit, Graph Convolutional Neural Network, Vulnerable Road Users

## I. INTRODUCTION

Humankind's social existence rely heavily on transportation mobility. Pedestrians, bikers, and motorcycle riders have traditionally been classified as vulnerable road users (VRUs) [1]. People with disabilities (particularly wheelchair users), the elderly, and children are usually more vulnerable than others when it comes to demographic qualities and walking ability. Paratransit is the most popular means of transportation for these VRUs. This group's most patronized paratransit schemes are still faced with the challenge of high-cost operations and low-quality service in transporting customers at the desired time from/to their desired origin/destination.

Paratransit is a public transportation system that can provide shared-ride door-to-door service with variable routes and times. Paratransit is one of the fastest-growing forms of transportation in the United States, with ridership set to double in the next few years [3]. Ridership has fueled the creation of models that can assist planners in designing new paratransit systems and expanding existing ones. In comparison to the traditional buses, it usually comprises of a Dial-A-Ride (DAR) system with a control center and a small vehicle fleet (generally 20 seats). When the control center receives transport requests, the buses, which have customizable routes and schedules, respond. Each customer shall provide details about where he/she is located, the time of pick-up or shipment, and the number of riders. In the control center, the dispatch man integrates customer information with vehicle location information and their attempting routes for preparing the new routes for vehicles using manual and automated dispatching techniques.

In the United States, most true dial-a-ride systems are highly subsidized, and the cost to the passenger can be very low. The bulk of operating expenses are spent on vehicle operations. Figure 1 shows patterns in US paratransit operating costs and fare revenues between 1995 and 2009. We can see that the rift between operating costs and fare revenue is gradually rising each year. In 2009, the gross operating costs for the paratransit services in the United States amounted to almost $5 billion, while fares received were $0.48 billion [5]. Of the overall operating costs, nearly $2.6 billion was spent on purchased transportation and $1.5 billion on vehicle operations.

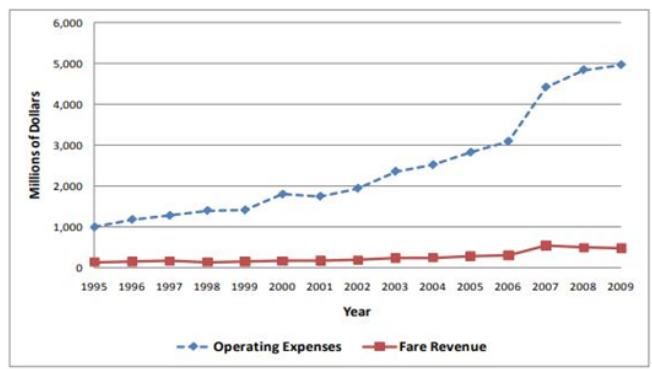

*Figure 1* Patterns in operating expenses and fare revenue for paratransit services in the US.
Source: America Public Transportation Association (APTA) database 2011 [5]

Kelvin Kwakye is a PhD candidate at the Industrial and Systems Engineering Department of the North Carolina Agricultural and Technical State University, 1601 E Market St, Greensboro, NC 27401, USA (email: kkkwakye@aggies.ncat.edu).

## II. RELATED WORK

Finding the optimal solution for a DAR problem which is Travelling Salesman Problem (TSP) is a nondeterministic polynomial time (NP-hard), even in the 2D Euclidean case where the nodes are 2D points and edge weights are Euclidean distances between pairs of points [6]. In practice, TSP solvers rely on carefully handcrafted heuristics to guide their search procedures for finding approximate solutions efficiently for graphs with thousands of nodes.

Jaw et al. [8] looked at multi-vehicle DARP with pick-up and drop-off windows, and in this case, vehicles cannot be idle while transporting passengers. Healy et al. suggested a new extension of the local search technique to solve the DARP in 1995 [9].

The majority of DARP research during the end of the 1990s focused on the development of contemporary heuristics or metaheuristics such simulated annealing, tabu search, and genetic algorithms (GA). Baugh et al. (1998) proposed clustering-based simulated annealing (SA) [10]. [11] devised a heuristic that uses a parallel insertion method to assign requests to routes before exchanging intra-route and inter-route data. The objective was to minimize the overall service expense of 276-312 requests for special cars for disabled people. Further advances were later achieved [11] through the introduction of a post-optimization tabu thresholding step after the parallel insertion step.

Today, state-of-the-art machine learning and deep learning techniques have been employed to solve TSP [2,7,12,13,14,17,19,20,21,22,23]. Another study [18] used a graph neural network estimator to estimate arrival time (ETA), which was implemented in Google Maps.

Among all these machine learning and deep learning techniques used in optimizing travel time, distance and costs estimation, the graph convolutional neural network algorithm has outperformed all these other algorithms. The current study justifies the adoption of graph convolutional neural network method on this basis.

## III. METHOD

### E. Problem Analysis

Let us consider n, a set of disabled persons who are in a different location and need paratransit rides to various destinations. Also, let N = {1, 2... n} be the set of disabled customers. In this study, we considered the number of paratransit vans having varying capacity denoted by the set Q to be sent to pick up locations L to get riders.

### B Assumptions

To model this problem as a vehicle routing problem, we will assume that;
1. The vans are only available at the depot, return to the depot, and only visit a disabled person once.
2. Also, it is worthy to note that all disabled persons at pick up points have different drop-off locations, say L different locations.

The VRP will be formulated using graphical theory and vehicle flow approach as shown in Figure 2 below.

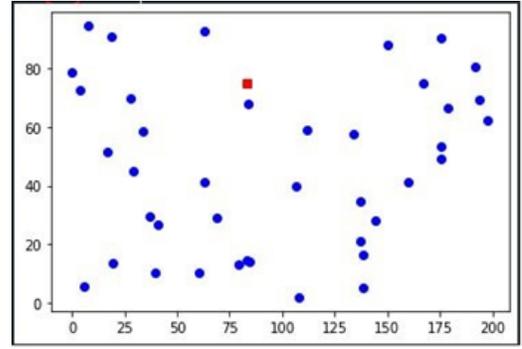

Fig *2 Nodes of the various pickup points in graphical form*
🔴 represents the depot
🔵 represent where the customers need to be picked up by a vehicle.

The objective of using VRP is to determine a set of minimal cost routes that satisfies all requirements and constraints given in the formulation in *Table 1* below.

*Table 1* Notations

| Notations | Definition |
|---|---|
| N | Set of challenged (Elderly/Disabled) people N= {1, 2, ..., n} |
| L | Set of locations (nodes)to be visited by the VRU person L = {1, 2, ..., l} |
| Q | Vehicle capacity |
| A | Set of arcs (i, j) for each pair of nodes i, j ε $L^2$ whereby i, j are locations 1 to 2 and i≠j |
| E | N U L U {0, n+l+1} Total number of nodes are all the locations they planned to visit, the starting point (depot), and the final destination. So, the starting and final destination at the depot are represented with 0 and n+l+1 respectively |
| $C_{ij}$ | Cost of moving from location i to j, such that i, j ε N, (l, j) ε E |
| $q_i$ | Number of people at a location l to be picked i ε N |

(Note: $q_o$ and $q_{n+l+1}$ are zeros because there is no demand or i.e. no one is to be picked at the depot or dropped-off at the depot).

### C. Declaration of Decision Variables

$X_{ij}$ ={1, if and only if there is a route that takes a person from i to j directly, for i, j ε N
　　 0, otherwise
$Y_j$ = Total number of people on the route that visits node j ε N up to this visit

### D. Model Formulation
DAR:
Objective Function

$$Min \sum_{i=0}^{n+l+1} \sum_{j=1}^{n+l+1} c_{ij} * x_{ij}$$

Equation (1)

Subject to:

$$\sum_{j=1, j\neq 1}^{n+l+1} x_{ij} = 1, i = 1,2,\ldots,n+l$$

Equation (2)

$$\sum_{i=0, i\neq h}^{n+l+1} x_{ih} - \sum_{j=1, j\neq h}^{n+l+1} x_{hj} = 0, h = 1,2,\ldots,n+l$$

Equation (3)

$$y_j \geq y_i + q_j x_{ij} - Q(1-x_{ij}), i,j = 0,1,\ldots,n+l+1$$

Equation (4)

$$q_i \leq y_i < Q$$

Equation (5)

$$\{x_{ij}\varepsilon\{0,1\}, i,j = 0,1,\ldots,n+l+1$$
$$\{y_i \geq 0, \quad i = 0,1,\ldots,n+l$$

Equation (6)

Equation (1) is the objective function which is aimed at minimizing the total travel cost.

Equation (2) is the constraint that ensures that each location (both the location of the challenged people (VRU) and their proposed destination are visited only once).

Equation (3) is the constraint that ensures that if a van arrives at a location, then it must leave to another, except when it is back at the depot.

Equation (4) and (5) together ensure that the vehicle capacity is not exceeded and avoid sub tours (i.e. ensure that all demands are served before eventually returning to the depot).

Equation (6) is for restrictions of the variables $X_{ij}$ (being binary) and $Y_j$

Let's consider riders who have requested for paratransit service in a day or two in advance, there is a need to pick these riders with vehicles with a certain capacity and use certain routes that will help reduce operating costs (i.e. to minimize travel distance). The operating scenarios considered in this study were;

- A randomized generated pick-up points (x, y) coordinates representing latitudes and longitudes in real-life case scenarios. Greensboro NC latitude and longitude were used 36.0726°, -79.7920° respectively.
- A generated cost using the Euclidean distances; that is the differences of the x-coordinates and the y-coordinates of each arc.
- A request scenario of 10, 15, 20, 30, 40 people
- Vehicle capacity of 10, 15, 20 seats

### F. Graph Neural Network

The graph convolutional network (GCN) has demonstrated exceptional performance on a variety of graph-based tasks, including node categorization and network representation.

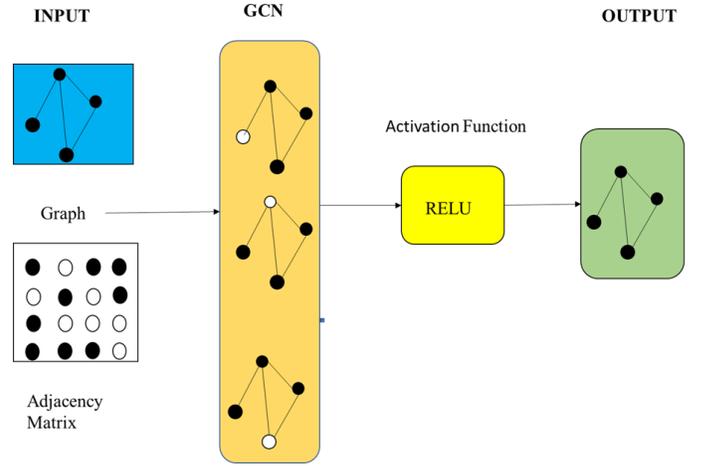

*Fig 4 Overview of the GCN approach*

The core notion of a graph-neural network is to collect information from each node and its neighbors, and graph neural networks have been shown to solve TSP problems []. The graph ConvNet model approach as shown in Fig 4 takes a 2D graph as input and produces an edge adjacency matrix that indicates the likelihood of edges occurring on the TSP tour with an optimal solution.

GCN Layer: σ(ÂXW)

σ: activation function =ReLU

X: input matrix ∈ $R^{N*d}$

W: parameter matrix ∈ $R^{d*d'}$

Graph Laplacian Â=Ď$^{1/2}$ĂĎ$^{-1/2}$, where Â = A+ I, Ď is a diagonal matrix with $Ď_{ii} = \sum_j Â_{ij}$

## IV. RESULTS

The study considered five different scenarios for the number of vulnerable users who will use the paratransit service. The scenarios considered were 10, 15, 20, 30, and 40 number of pick-up requests by VRUs at a time as shown in Table 2. Also, vehicles with seat capacities of 10, 15, and 20 were considered.

Table 2 summarizes the results from simulating the different case scenarios.

Table 2. Simulation Results

| Number of VRUs n | Capacity Q | Nodes Explored | Simplex Iterations | Run Time | Objective Cost (Distance) | Gap % |
|---|---|---|---|---|---|---|
| 10 | 10 | 1 | 33 | 0.04 | 0.0112255 | 0.0% |
| 15 | 10 | 1 | 103 | 0.11 | 0.0200046 | 0.0% |
| 20 | 10 | 23 | 498 | 0.56 | 0.01795 | 0.0% |
| 30 | 10 | 1 | 312 | 0.31 | 0.0282072 | 0.0% |
| 40 | 10 | 1 | 1568 | 1.97 | 0.050841 | 0.0% |
| 10 | 15 | 22 | 326 | 0.27 | 0.0130656 | 0.0% |
| 15 | 15 | 6761 | 117645 | 5.63 | 0.0130461 | 0.0% |
| 20 | 15 | 2832 | 35141 | 4.12 | 0.0181658 | 0.0% |
| 30 | 15 | 7109 | 161176 | 30.04 | 0.0275969 | 3.4743% |
| 40 | 15 | 813 | 13611 | 30.03 | 0.0496514 | 9.2159% |
| 10 | 20 | 478 | 6672 | 0.41 | 0.0134435 | 0.0% |
| 15 | 20 | 22749 | 425665 | 20.55 | 0.012733 | 19.1248% |
| 20 | 20 | 6750 | 290342 | 30.02 | 0.019809 | 38.1247% |
| 30 | 20 | 787 | 28424 | 30.02 | 0.0229079 | 26.2459% |
| 40 | 20 | 777 | 36870 | 30.03 | 0.0301532 | 41.3090% |

to assign to pick up customers and informs the drivers of the routes to take, hence reducing the cost of operations.

### A. Graph of Variations

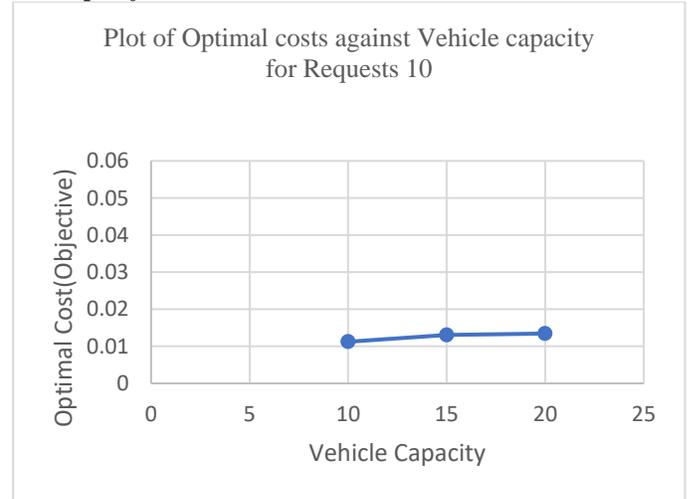

Figure 6 Graph showing the trend of optimal cost versus vehicle capacity for the number of VRU request 10

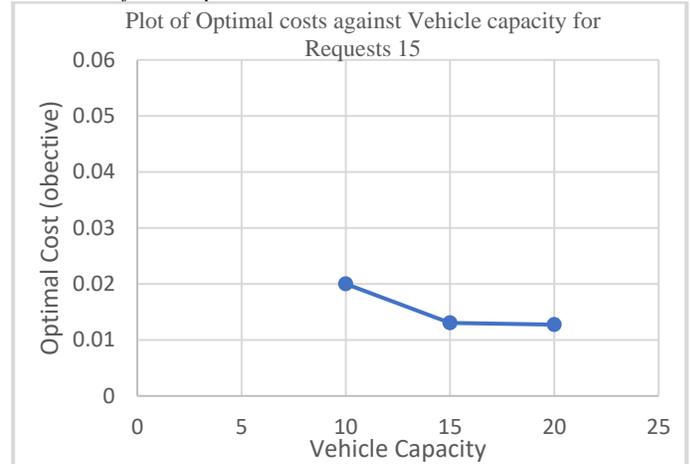

Figure 7 Graph showing the trend of optimal cost versus vehicle capacity for the number of VRU request 15

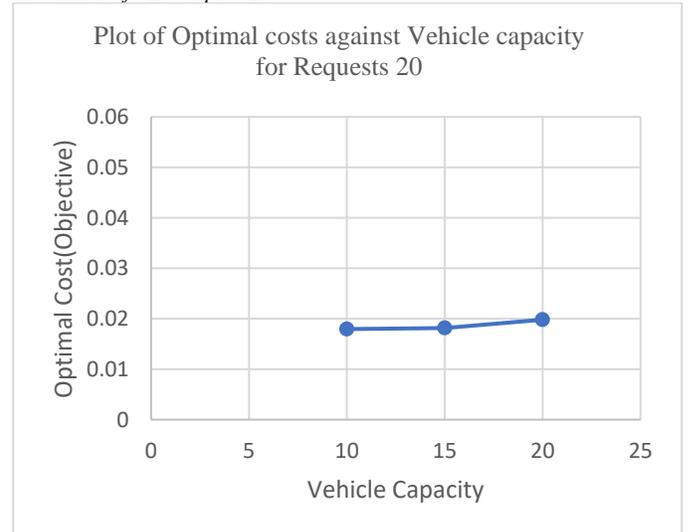

Figure 8 Graph showing the trend of optimal cost versus vehicle capacity for the number of VRU request 20

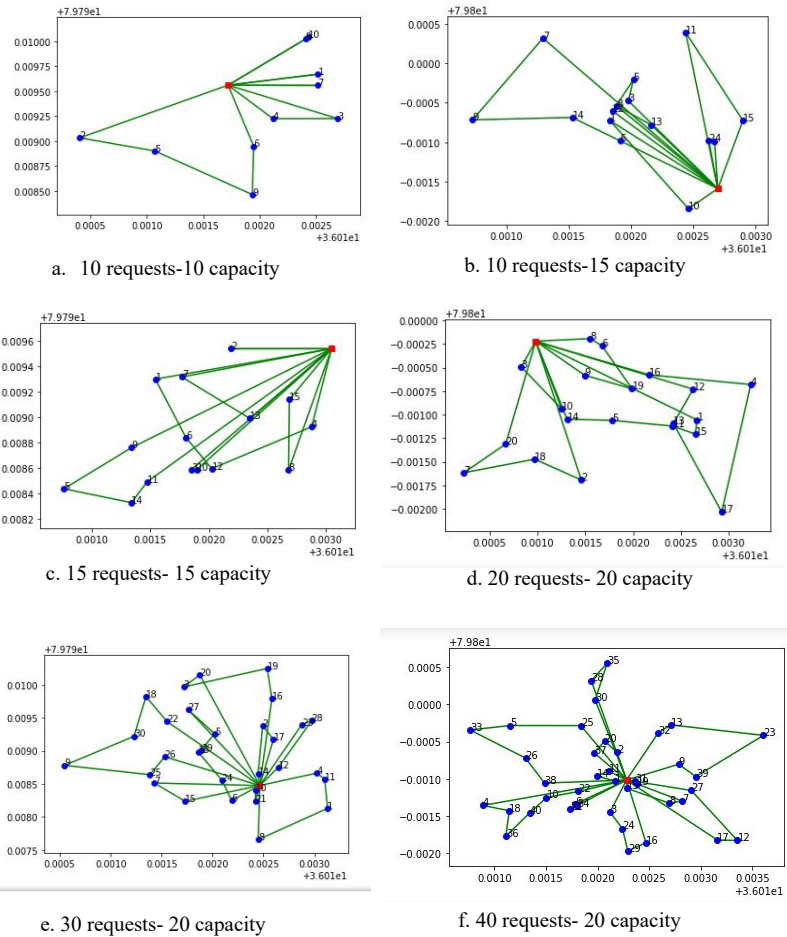

a. 10 requests-10 capacity
b. 10 requests-15 capacity
c. 15 requests- 15 capacity
d. 20 requests- 20 capacity
e. 30 requests- 20 capacity
f. 40 requests- 20 capacity

Fig 5. optimized routing plan after running the different scenarios

For example, *Figure 5 (a)* is an optimized routing plan for a scenario of 10 requests with a vehicle capacity of 10. The red point represents the location of the depot and the blue markers are points where requests were made on a map in a graphical representation. The algorithm solves the problem by optimizing the routes into three (3) to help serve the demands of the 10 vulnerable persons who requested the rides. This, therefore, helps the operators to know the number of vehicles

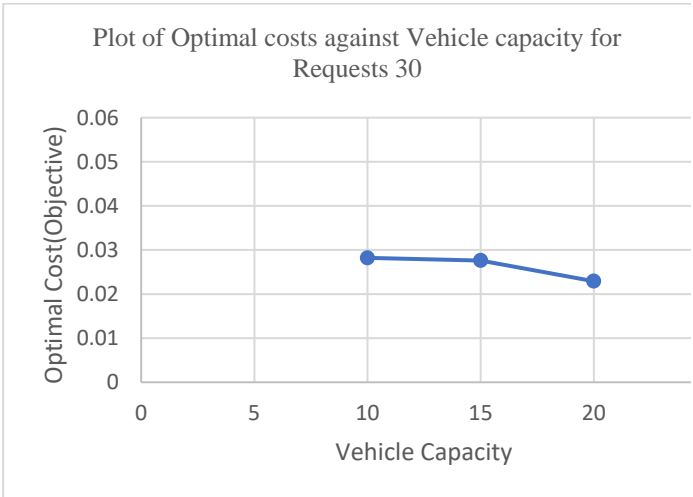

*Figure 9 Graph showing the trend of optimal cost versus vehicle capacity for the number of VRU request 30*

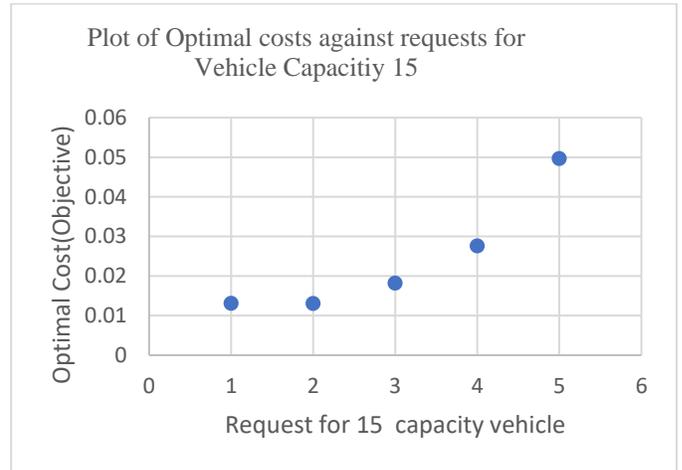

*Figure 12 Graph showing the trend of optimal cost versus the number of VRU request for vehicle capacity 15*

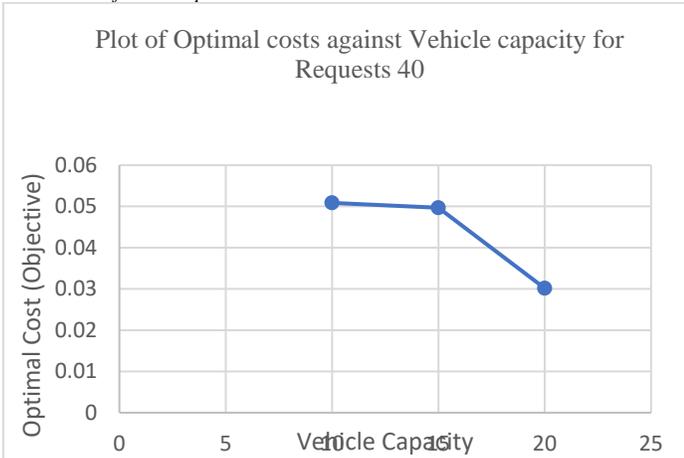

*Figure 10 Graph showing the trend of optimal cost versus vehicle capacity for the number of VRU request 40*

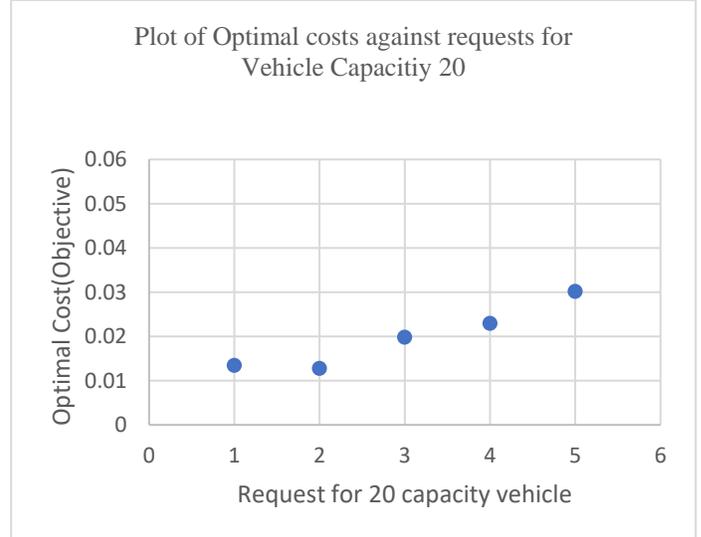

*Figure 13 Graph showing the trend of optimal cost versus the number of VRU request for vehicle capacity 10*

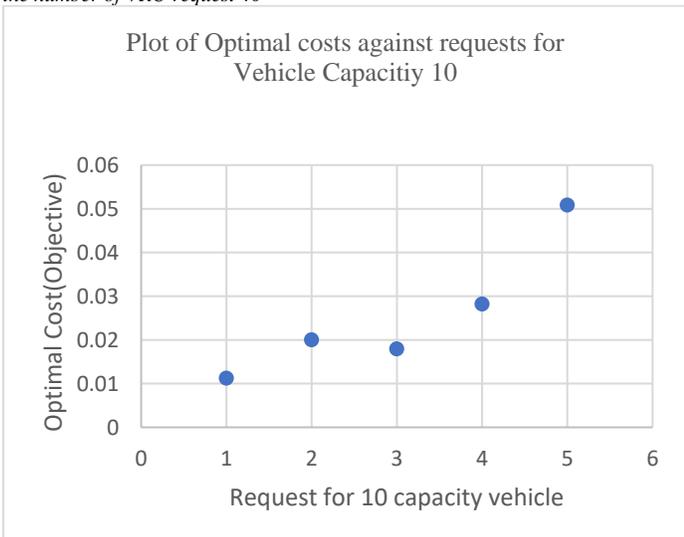

*Figure 11 Graph showing the trend of optimal cost versus the number of VRU request for vehicle capacity 10*

*Figures 11-13* are plots of the optimal cost against requests for vehicle capacities of 10, 15, and 20, respectively. The x-axis numbered 1-5 represents requests 10, 15, 20, 30, 40, respectively. It can be found from these graphs that the optimal cost for a vehicle with a capacity of 20 for a request of 40 was 0.030153, while the optimal cost for a vehicle capacity of 15 and 10 for the same request of 40 is 0.04965 and 0.05085 respectively. This means that the optimal cost decreases with increasing capacity for a requested scenario of 40. This, therefore, implies, that it would be more economical for operators to use a vehicle with a capacity of 20 when the number of requests is 40. Also, in the request scenarios of 30, The optimal cost was decreasing again with increasing vehicle capacities.

For request scenarios of 20, the optimal cost can be seen to be increasing with a rise in vehicle capacities. The minimum cost of a vehicle with a capacity of 10 seats is 0.01795 compared to a capacity of 15 and 20. The same is true of a requested scenario of 10 with a vehicle capacity of 10 seats, which is highly recommended for such a scenario. In scenarios of 15 requests, the vehicle with a seating capacity of 10 was costly as compared to 15 and 20.

## V. Limitations and Future Works

The study's shortcomings were as follows: Insufficient data on paratransit operations, such as a service area, hot spot areas, fleet size (vehicles), cost, demand density (trips/miles-day), peak-hour demand (trips/hr.), average vehicle speed (mi/hr.), vehicle capacity (passengers), pickup time window (min), available stop IDs, trip IDs, etc. To aid the model in dynamic scenarios, the maximum travel time ratio, maximum route duration (min), and passenger boarding time (min) will be used.

A model for dynamic situations and huge case scenarios would be developed in future research.